\documentclass[letterpaper, 10 pt, conference]{ieeeconf}
\IEEEoverridecommandlockouts
\usepackage{fdsymbol}
\usepackage{multirow}
\usepackage{adjustbox}
\usepackage{hyperref}
\usepackage{graphicx,subcaption}

\usepackage[nolist]{acronym}
\begin{acronym}
\acro{vo}[VO]{Visual Odometry}
\acro{cnn}[CNN]{Convolutional Neural Network}
\acro{rnn}[RNN]{Recurrent Neural Network}
\acro{mdn}[MDN]{Mixture Density Network}
\acro{mse}[MSE]{Mean Squared Error}
\acro{aft}[AFT]{Asynchronous Fusion Transformer}
\acro{aftvo}[AFT-VO]{\acs{aft}-VO}
\acro{lstm}[LSTM]{Long Short-Term Memory}
\acro{mlp}[MLP]{Multilayer Perceptron}
\acro{imu}[IMU]{Inertial Measurement Unit}
\acro{rpe}[RPE]{Relative Pose Error}
\acro{ekf}[EKF]{Extended Kalman Filter}
\end{acronym}

\title{\LARGE \bf
\acs{aftvo}: Asynchronous Fusion Transformers for \\Multi-View Visual Odometry Estimation
}

\author{Nimet Kaygusuz, Oscar Mendez, Richard Bowden$^{1}$%
\thanks{$^{1}$All authors are with the University of Surrey, UK
        {\tt\small \{n.kaygusuz, o.mendez, r.bowden\}@surrey.ac.uk}}%
}

\def\eg{\emph{e.g. }}
\def\ie{\emph{i.e. }}
\def\etal{\emph{et al. }}

\begin{document}

\maketitle
\thispagestyle{empty}
\pagestyle{empty}

\begin{abstract}

Motion estimation approaches typically employ sensor fusion techniques, such as the Kalman Filter, to handle individual sensor failures. 
More recently, deep learning-based fusion approaches have been proposed, increasing the performance and requiring less model-specific implementations.
However, current deep fusion approaches often assume that sensors are synchronised, which is not always practical, especially for low-cost hardware. To address this limitation, in this work, we propose \acs{aftvo}, a novel transformer-based sensor fusion architecture to estimate \acs{vo} from multiple sensors. Our framework combines predictions from asynchronous multi-view cameras and accounts for the time discrepancies of measurements coming from different sources.

Our approach first employs a \acf{mdn} to estimate the probability distributions of the 6-DoF poses for every camera in the system.
Then a novel transformer-based fusion module, \acs{aftvo}, is introduced, which combines these asynchronous pose estimations, along with their confidences. More specifically, we introduce \textit{Discretiser} and \textit{Source Encoding} techniques which enable the fusion of multi-source asynchronous signals.

We evaluate our approach on the popular \emph{nuScenes} and \emph{KITTI} datasets.
Our experiments demonstrate that multi-view fusion for \acs{vo} estimation provides robust and accurate trajectories, outperforming the state of the art in both challenging weather and lighting conditions.

\end{abstract}

\section{INTRODUCTION}

\label{sec:intro}

\acf{vo} can be described as the process of estimating the relative pose and motion of a moving camera given a series of images. Traditional approaches have been studied for decades and are used to extract motion in applications ranging from robotics to autonomous driving. They generally follow the standard procedure of feature detection followed by robust matching and use projective geometry to estimate camera motion \& therefore odometry. However, these approaches are not robust against poor lighting conditions and low textured environments, where the feature detection and matching tend to fail. In recent years, learning-based \ac{vo} approaches have become popular and achieved competitive performance \cite{wang2017deepvo, li2018undeepvo, xue2019beyond} due to their ability to extract high-level representations without relying on hand-crafted features. 

Most of the current \ac{vo} approaches employ a single monocular or stereo camera. However, \ac{vo} is prone to failure when the quality of the captured image is degraded by occlusion, lighting change, etc.
Various approaches have been proposed to address this issue by combining cameras with non-visual sensors such as IMUs \cite{clark2017vinet, chen2019selective}. They demonstrate improved performance when camera motion estimation fails. However, employing multiple cameras is a clear way to provide robustness to individual camera failures. This has led to employing multiple cameras with different views as an alternative way to provide robustness to individual camera failures \cite{tribou2015multi}, \cite{paudel20192d}.

Multi-camera systems \cite{liu2018towards, tribou2015multi}, follow an optimisation-based approach, however, they assume that the sensors are time-synchronised. Unfortunately, synchronous multi-source signals are hard to acquire. This is because obtaining images from multiple cameras at the same time requires `gen-locked' or synchronised hardware and sufficient bandwidth to capture and process those images with the associated cost implication. Thus new advancements are needed that can enable the use of low-cost asynchronous sensors.

Motivated by the limitations of the current multi-camera approaches, in this work, we aim to estimate robust \ac{vo} using images obtained from multiple unsynchronised cameras. To achieve this, we introduce a novel transformer-based deep fusion framework. Our approach can fuse signals from any number of sources and, more importantly, it does not require those sources to be synchronised or to be at the same frequency.

As inputs, our approach takes variable-length video sequences from any number of cameras attached to a vehicle. We first process every video stream separately and estimate independent motion and uncertainties using a deep \ac{mdn}-based \ac{vo} module \cite{kaygusuz2021iros}. Due to the asynchronous nature of these signals, naive fusion approaches, such as concatenating representations corresponding to the same time step, are not applicable. To address this, we introduce a novel \textit{Discretiser} module which enables our system to positionally encode continuous timestamps. 
Thus, we also introduce \textit{Source Encoding} which enables the fusion module to differentiate signals coming from different cameras as predictions from different cameras do not have any indication of their origin. 

The contributions of this paper can be summarised as:
\begin{itemize}
    \item We propose \acs{aftvo}, a transformer-based deep fusion framework which can fuse asynchronous signals from multiple sources. We apply our fusion approach to the multi-view \ac{vo} estimation problem.
    \item We introduce two new types of encoding approach into transformers: (1) \textit{Discretiser}: to represent the temporal relationship between asynchronous time series (2) \textit{Source Encoding}: to inform the model about the source of the information.
    \item We evaluate our approach on two popular public datasets, namely, KITTI and the nuScenes datasets, and achieve state-of-the-art performance.
\end{itemize}

\section{Related Work}
\label{sec:related-work}

Traditional \ac{vo} approaches can be split into two categories which are direct and feature-based methods. Direct methods estimate the motion by minimising the photometric error \cite{engel2017direct, newcombe2011dtam}. Feature-based methods estimate the motion by extracting and matching a set of feature points on consecutive frames \cite{mur2015orb, nister2004visual, yang2021asynchronous}. While direct methods perform well in low texture settings, feature-based methods are more successful at higher velocities.

In recent years, deep learning-based \ac{vo} methods achieved promising results. These methods learn to extract high-level features from the data and therefore do not need hand-crafted features. Current learning-based \ac{vo} approaches can be classified as supervised, which we employ in this work, or unsupervised. One of the earliest supervised approaches was proposed by Mohanty \etal \cite{mohanty2016deepvo} who estimate \ac{vo} using a \ac{cnn}. Wang \etal \cite{wang2017deepvo} combine a \ac{cnn} with a \ac{rnn} to enhance temporal dependencies. Kaygusuz \etal \cite{kaygusuz2021iros} utilise an \ac{mdn} to estimate \ac{vo} and its uncertainty. Saputra \etal \cite{saputra2019learning} introduce geometric loss constraints to increase generalisation capability. Xue \etal \cite{xue2019beyond} propose a memory module to store global information which is then used to refine the estimated poses. 
To reduce the dependency of labelled data, unsupervised methods have also been studied. Zhou \etal \cite{zhou2017unsupervised} proposed learning depth and motion jointly. Li \etal \cite{li2018undeepvo} and Zhan \etal \cite{zhan2018unsupervised} use stereo images to recover the scaled \ac{vo}.
While unsupervised methods are more easily scaled, to date they have not performed as well as supervised methods. 
Additionally, both supervised and unsupervised methods rely on single-camera models, which are liable to fail in difficult environmental conditions (\eg glare, low texture, etc). 

To overcome motion estimation failure, researchers have studied sensor fusion techniques.
Geneva \etal \cite{geneva2018asynchronous} propose a multi-sensor fusion system based on a factor graph-based optimisation framework.
Sola \etal \cite{sola2008fusing} employed a Kalman Filter to fuse measurements from a multi-camera system. Liu \etal \cite{liu2018towards} used multiple cameras to build a classical \ac{vo} approach. Zhang \etal \cite{zhang2018vins} studied a multi-view visual-inertial odometry system. 

In recent years, learning-based fusion for motion estimation has been studied. Kaygusuz \etal \cite{kaygusuz2021itsc} introduced a learning-based multi-view fusion approach in which they assume that the cameras are synchronised. Clark \etal \cite{clark2017vinet} proposed a learning-based approach to visual-inertial odometry by simply concatenating features from a camera and an inertial sensor. Chen \etal \cite{chen2019selective} also studied visual-inertial odometry and introduced a selective fusion approach. However, they relied on only inertial measurements when the camera images are degraded by occlusion, sunlight etc. Contrary to previous techniques, in this work, we propose the first learning-based asynchronous multi-view \ac{vo} system.

Historically, deep learning approaches have used recurrent architectures to model time series. 
More recently, transformer models \cite{vaswani2017attention} have been introduced for time series data, in the domain of machine translation. They have drastically increased the performance of sequence-to-sequence tasks such as, speech recognition \cite{irie2019language}, language understanding \cite{devlin2019bert} and sign language translation \cite{camgoz2020sign}. 

Inspired by the recent success of transformers, in this work we propose a novel architecture to fuse measurements for the estimation of \ac{vo}. We formulate our fusion approach as a sequence-to-sequence problem where the inputs are asynchronous and multi-source and do not necessarily have the same cardinality as the output. 

\begin{figure*}[t]
\begin{center}
   \includegraphics[width=0.80\linewidth]{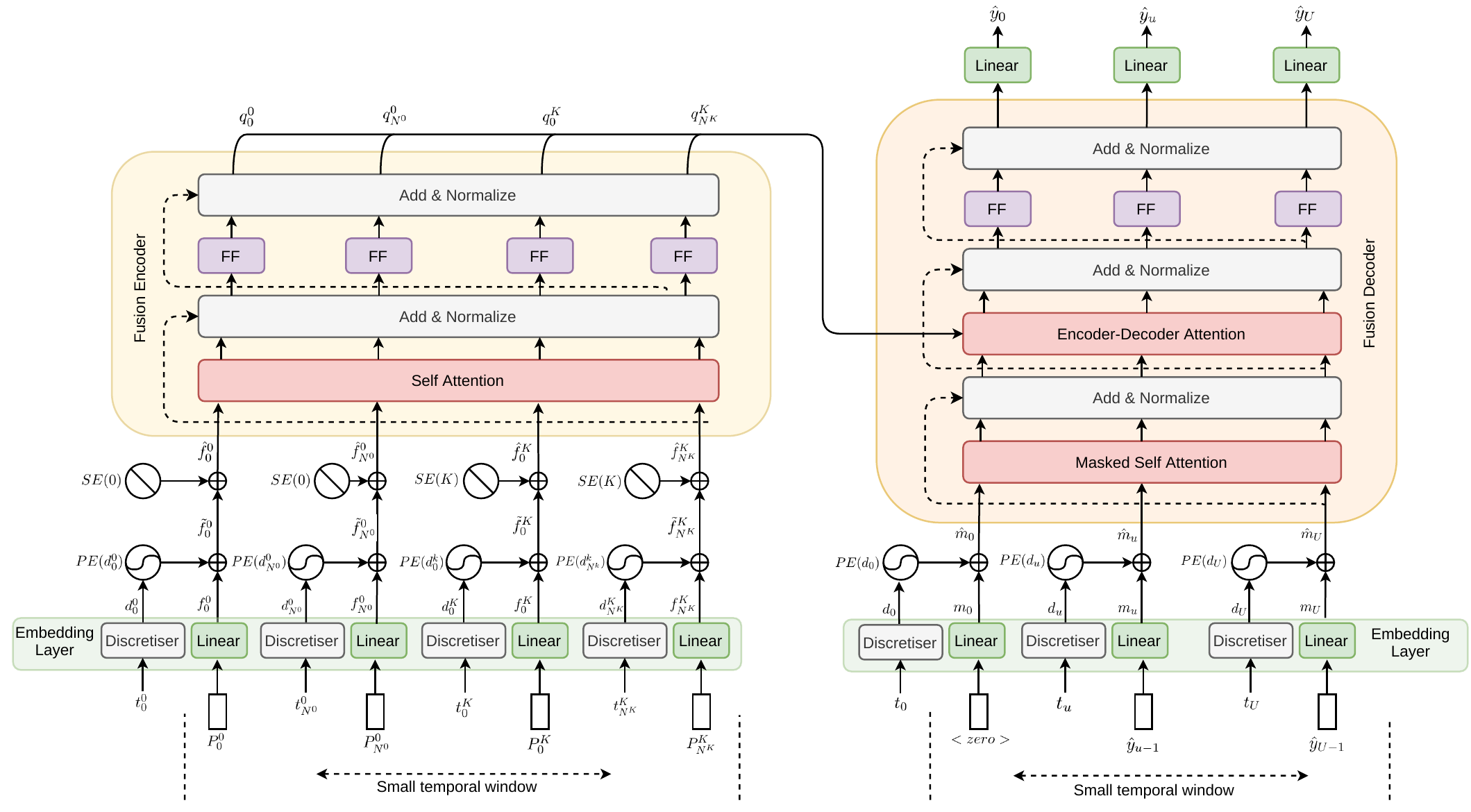}
\end{center}
   \caption{Overview of the proposed asynchronous fusion module. To fuse asynchronous information and differentiate signals that are coming from different cameras, we utilise two novel encoding approaches, \emph{Discretiser} and \emph{Source Encoding (SE)}.}
\vspace{-0.40cm}
\label{fig:architecture}
\end{figure*}

\section{Multi-View Visual Odometry}
\label{sec:methodology}

We study multi-view \ac{vo} as a fusion problem with observations captured by different sensors at different timestamps. Our task is to estimate vehicle trajectory, $Y_{1:U}$, given variable-length asynchronous video sequences, $\mathcal{V} = \{\mathcal{V}^0,...,\mathcal{V}^K\}$ from $(K+1)$ cameras all rigidly mounted to the same vehicle.
A video stream coming from the $k^{th}$ camera, $C^k$, can be written as:
\begin{equation}
    \mathcal{V}^k = \{(I_0^k, t_0^k),...,(I_{n}^k, t_{n}^k),...,(I_{N^k}^k, t_{N^k}^k)\} 
\end{equation}
where $I_n^k$ represent the $n^{th}$ video frame captured at the timestamp $t_n^k$, and $N^k$ represents the total number of captured frames by the $k^{th}$ camera. Note that due to asynchronicity, frames coming from different cameras, $I_n^i$ and $I_n^j$, that share the same order index, $n$, do not necessarily share the same timestamp, \ie $t_n^i \neq t_n^j$.

Our network can be broken into two sections: a \acf{mdn} and our novel \acf{aft}. For each camera, we first estimate a mixture of Gaussian distributions of the 6-DoF relative poses using an \ac{mdn}. We then fuse all pose distributions estimated through different arms of the network in the \ac{aft} module.

\subsection{\acf{mdn}}
\label{sec:mdn}
Learning-based approaches to \ac{vo} generally regress a 6-DoF pose directly. In this work, we employ an \ac{mdn} similar to \cite{kaygusuz2021iros} and estimate the probability distribution of the 6-DoF pose, conditioned on the input images. Estimating the pose as a mixture of distributions provides our fusion module with an indicator of the estimation uncertainty.

We first extract latent representations from consecutive image pairs in each camera stream using a \ac{cnn}. We employ FlowNet \cite{dosovitskiy2015flownet} as our \ac{cnn} backbone which has proven to be successful for learning features that are geometrically meaningful. We formalise this operation for two consecutive images, $(I_{t-1}^k, I_t^k)$, from camera $C^k$ at time $t$ as:\looseness=-1
\begin{equation}
    w_t^k = \mathrm{CNN}([I_{t-1}^k, I_t^k]).
\end{equation}
where $[]$ operation represents the channel-wise concatenation of consecutive images.

Motion estimation is highly dependant on temporal information. Thus, we employ an \ac{rnn}, which takes in the latent representation, $w_t^k$, at time $t$ to produce temporally enriched representations, $r_t^k$, as:
\begin{equation}
    r_t^k = \mathrm{RNN}(w_t^k,h_{t-1}^k).
\end{equation}
where $h_{t-1}^k$ represents the \ac{rnn}'s previous hidden state. 

We then pass the temporally enhanced representations through the \ac{mdn} module. We build our mixture model as a multivariant Gaussian distribution and formalise a mixture model, $P_t^k$ as:
\begin{equation}
{P}_t^k = \{(\alpha^1_t\mathcal{N}_t(\mu^1, \sigma^1))^k, ..., (\alpha^X_t \mathcal{N}_t(\mu^X, \sigma^X))^k\}
\label{eq:mixparams}
\end{equation}
where $X$ denotes the number of mixture components, while $\mu$, $\sigma$ and $\alpha$ represent the mean, standard deviation and mixture coefficient, respectively. A linear combination of mixture components produces the probability density of the target pose, $y_{(t-1,t)}^k$, and can be written as:
\begin{equation}
\vspace{-0.05cm}
    p\left(y_{(t-1,t)}^k|r_t^k\right) = \sum_{i=1}^{X}
    \left(\alpha_{i}\left(r_t\right)\right)^k \phi_i \left(y_{(t-1,t)}^k|r_t^k\right)
\end{equation}
where $\phi_i$ is the conditional density function for the $i^{th}$ component. The mixture coefficient, $(\alpha_i\left(r_t\right))^k$, represents the probability of the target pose, $y_{(t-1,t)}^k$, being generated by the $i^{th}$ component. We train our \ac{mdn} module by minimising the negative log likelihood of the ground truth poses coming from our estimated mixture model.

\subsection{\acf{aft}}
\label{sec:aft}

We consider the task of multi-view \ac{vo} estimation as a sequence-to-sequence learning problem and propose a novel transformer-based approach to fuse complementary information coming from multiple asynchronous cameras. 

The input to the asynchronous fusion module is the pose and uncertainty estimates from the \ac{mdn} module with their corresponding timestamps, $\{(P_0^k, t_0^k),...,(P_n^k, t_n^k),...,(P_{N^k}^k, t_{N^k}^k)\}$. 
One thing to note is that due to asynchronicity, estimates coming from different cameras within a selected time interval, \eg $C^i$ and $C^j$, will not necessarily share the same cardinality, \ie $\medlozenge(N^i \neq N^j)$, or timestamps of the same time indices, $n$, will not necessarily be equal, \ie $\lozenge (t_n^i \neq t_n^j)$. We first project the predictions from individual cameras into a latent representation using a linear layer as:
\begin{equation}
    f_n^k = \mathrm{Linear}(P_n^k); 
\end{equation}
where $P_n^k$ is the pose estimate and its uncertainty (See Section~\ref{sec:mdn}) from the $k^{th}$ camera at the $n^{th}$ time step.

Transformers do not inherently contain any temporal modelling, and function by utilising attention mechanisms. To this end, positional encoding techniques are used to inject the order of the inputs by creating a unique vector for each discrete step.
Although positional encoding has been applied to sequence-to-sequence learning problems, such as text-to-text \cite{rosendahl2019analysis, vaswani2017attention} and video-to-text tasks \cite{akbari2021vatt}, they assume the consecutive items to be equidistant in the time domain \eg the distance between every consecutive input is equal. This is a valid assumption for tasks where input comes from a single source with a fixed frequency. However, this is not applicable to our problem as the information comes from multiple asynchronous sources where the inputs are not equally distant from each other. To represent continuous time information and to differentiate signals coming from different sources, we propose two novel encoding approaches, namely \emph{Discretiser} and \emph{Source Encoding}.

\textbf{Discretiser:}
In this module, our goal is to represent the continuous timing information with a set of discrete equidistant time step bins. Although the timings of captured images are generally reported in integer format, \ie microseconds, the cardinality of this space is much larger than what transformers are designed to work with, \ie number of words in a sentence. Also, the information change in this time domain is much slower than the text domain. For example, information gained from $1$ microsecond of a $5$ seconds video is much smaller than what is gained from a word in a sentence with a cardinality of $10$. Furthermore, the frequency of sensors can be varied throughout a capture, resulting in non-equidistant frames. 

To address this issue we propose to discretise the continuous time domain into bins. Given a sequence of timestamps from an arbitrary time window, we first normalise them by finding the earliest stamp and subtracting it from all others. 
We then divide the time axis into smaller chunks, $Z$, and group the measurements into bins that have been captured close in time. Chunk size, $Z$, affects the discretisation sensitivity. In our experiments we choose 20ms as the chunk size, however reducing it does not affect the performance significantly. We formalise this operation as:
\begin{equation}
    d_n^k = \mathrm{round}(\frac{t_n^k -  \mathrm{min}(\{t_0^0,...t_{N^0}^0\},...,\{t_0^K,...t_{N^K}^K\})}{Z})
\end{equation}
where $d_n^k$ represents the discretised form, \ie the bin index, of the timestamp $t_n^k$, and $Z$ is the quantisation step size. Note that the $d_n^k$ is rounded down to the nearest integer. This allows us to convert the time axis into fewer discrete units, where each integer represents a bin of timestamps.

Next, we use these discretised representations to obtain relative position embeddings for inputs in a sequence. We employ the positional encoding approach proposed in \cite{vaswani2017attention}, which uses a dense sequential set of integers, \ie the position of the word in a sentence, to index the positional encoding matrix. However, our discretisation based use of positional encoding works with a sparse set of integers, \ie the bin indexes of timestamps. This allows our model to learn the temporal relationship between asynchronous multi-source signals that lie on the continuous time axis. 
We achieve this by using $d_n^k$ as an index to select from the positional encoding matrix. We formalise positional encoding of the inputs as:
\begin{equation}
    \Tilde{f_n^k} = f_n^k + \mathrm{PositionalEncoding}(\mathrm{Discretiser}(t_n^k))
\end{equation}

\textbf{Source Encoding:}
Transformer architectures do not have any built-in way to identify the source of signals. However, our model needs to be able to differentiate between sources so that it can exploit the complementary information coming from them. To achieve this we propose \emph{Source Encoding}. For each source in our system, we generate a unique one-hot vector.
We then pass these one-hot representations through a linear layer that is learnt during training to obtain unique source encoding for each camera. We then combine this additional source information into the positionally encoded inputs as: 
\begin{equation}
    \hat{f_n^k} = \Tilde{f_n^k} + \mathrm{SourceEncoding(k)}
\end{equation}

\textbf{Fusion Encoder:}
After embedding the relative temporal position and the camera source information into the input representations, we fuse them using a transformer encoder model. The inputs to the encoder are first modelled by a self-attention layer which learns the correlations between different timestamps. The outputs of the self-attention layer are then passed through a position-wise feed forward layer. Residual connections and layer normalisation are added after every operation. We formulate the fusion encoding process as:
\begin{equation}
    q_n^k = \mathrm{FusionEncoder}\left(\hat{f}_n^k | \{\hat{f}_{1:N^0}^0 ... \hat{f}_{1:N^K}^K\}\right)
\end{equation}
where $q_n^k$ denotes the fused representation corresponding to the $n^{th}$ frame coming from the $k^{th}$ camera. 
Note that to produce $q_n^k$, the encoder attends to the representations of all predictions from all available sources in an arbitrarily chosen time window.

\textbf{Fusion Decoder:} The encoder produces a set of fused representations, $Q = \{ q_n^k \forall n,k \}$, within a time window. We use these latent representations, $Q$, to decode the vehicle's odometry in an autoregressive manner.

During training, we concatenate a zero vector, $\textless0\textgreater$, to the beginning of the target pose sequence, $\{(\textless0\textgreater,t_0),...,(\hat{y}_{u-1}, t_u),...,(\hat{y}_{U-1}, t_{U})\}$, which is then used as the input of the decoder. This operation is analogous to the addition of the special beginning of sentence token in symbolic sequence-to-sequence tasks, which allow teacher forcing to be applied during training. We inject positional encoding using the \emph{Discretiser} similar to the fusion encoder module. These embedded representations, $\hat{m}_u$, are then passed to a masked self-attention layer. Unlike the self-attention utilised in the encoder module, masking prevents the previous time steps from gaining information from the future while estimating a pose. This limitation is key, as it allows the system to function at inference time when the information is not available to the model.
We then combine the output of the fusion encoder and decoder self-attention. These representations are passed to an encoder-decoder attention module. 
This module learns the relationship between source and target pose sequences. As with the fusion encoder, all layers are followed by residual connections and layer normalisation. We formulate this fusion decoder process as:
\begin{equation}
    \hat{y}_{(t-1,t)} = \mathrm{FusionDecoder}\left(\hat{m}_u|\hat{m}_{1:u-1},\{q_{1:N^0}^0 ... q_{1:N^K}^K\} \right)
\end{equation}
During inference, we query the decoder with timestamps for which we want a pose estimate. In our experiments, we used the ground truth timestamps that are available to decode the vehicle's odometry, but the model can handle arbitrary timestamps. We train our asynchronous fusion transformer model using a \ac{mse} loss function as:
\begin{equation}
    \mathcal{E} = \frac{1}{U} \sum^U_{t=1}{||y^\tau_{(t-1, t)} - \hat{y}^\tau_{(t-1, t)}||^2_2} + \Omega {||y^\psi_{(t-1, t)} - \hat{y}^\psi_{(t-1, t)}||^2_2}
\end{equation}
where $y^\tau_{(t-1, t)}$ and $y^\psi_{(t-1, t)}$ represents the translation and rotation of the 6-DoF ground truth poses, respectively. $\Omega$ is a parameter to increase the importance of rotation errors ($100$ in our experiments) and $U$ is the number of time steps within a given arbitrary time interval.

\section{Experiments}
\label{sec:experiments}
In this section, we evaluate the proposed \acs{aftvo} approach and compare it against the single and multi-sensor state-of-the-art \ac{vo} techniques.

\subsection{Dataset \& Implementation Details}
We conduct our experiments using two popular autonomous driving datasets, namely KITTI \cite{geiger2013vision} and nuScenes \cite{caesar2020nuscenes}. KITTI has become the benchmark dataset for motion estimation. However, it is not an ideal dataset for asynchronous multi-camera fusion systems, considering it only has a stereo camera pair. For completeness, we evaluate our model on KITTI to demonstrate that our system can also work with synchronous cameras. In addition, the nuScenes dataset has $6$ asynchronous cameras giving a $360$ degree view, which makes it an ideal dataset to evaluate our approach.

We implement our model using the PyTorch deep learning framework. Pre-trained FlowNet weights are used for our \ac{cnn} initialisation \cite{dosovitskiy2015flownet}. We build our transformers with 4 layers for both encoder and decoder, each layer containing $512$ hidden units and 4 heads. The network parameters are initialised with the Xavier approach. We train our model using the Adam optimiser $(\beta_1=0.9, \beta_2=0.999)$ with a batch size of $32$ and a learning rate of $0.0005$. To assess the performance of our approach we use the standard \ac{rpe} metric from the \textit{evo} evaluation toolkit \cite{grupp2017evo}. 

\subsection{Ablation Studies}
In our first set of experiments, we ablate input sources and different components of the \acs{aftvo}. We conduct these experiments on the nuScenes dataset. We perform two sets of ablation studies to measure (1) the effect of fusing different sets of cameras, (2) the benefits of the proposed modules.

\textbf{Camera Ablation:}
In our first ablation study, we compare the effect of fusing different sets of cameras with complementary views.

Table \ref{table:cam_ablation} shows fusion results with different sets of cameras. As can be seen, the model that utilises a single view, namely front camera (F), has the largest error. Fusing other view combinations such as front and back cameras (F+B), and three front cameras (F+FL+FR), improves the performance. One interesting finding is that the two-camera system (F+B), outperforms the three-camera system, (F+FR+FL). We believe this is due to the significant increase in view coverage between these two setups. For example, there is natural overlap between cameras F, FR and FL resulting in considerable redundancy whereas (F+B) provides more complementary information. 

Ultimately, we see that fusing all the cameras yields the best results, further suggesting that the proposed approach benefits from the complementary information coming from different views. 

\begin{table}[!h]
\centering
\caption{Average RPE (m) results from the camera ablation experiments. (F: Front, FR: Front Right, FL: Front Left, B: Back, BR: Back Right, BL: Back Left)} 
\begin{adjustbox}{width=0.95\linewidth}
\begin{tabular}{l||ccc}
Fused Camera & {RMSE} & {Max} & Mean $\pm$ std \\ \hline
F & 0.05482 & 0.17238 & 0.04393 $\pm$ 0.03190\\
F+FR+FL & 0.05226 & 0.17684 & 0.04157 $\pm$ 0.03102\\
F+BR+BL & 0.04870 & 0.16094 & 0.03889 $\pm$ 0.02899\\
F+B & 0.04853 & 0.16183 & 0.03864 $\pm$ 0.02867\\
F+FR+FL+B & 0.04608 & 0.15116 & 0.03696 $\pm$ 0.02682\\
All & \textbf{0.04529} & \textbf{0.14983} & \textbf{0.03637} $\pm$ 
\textbf{0.02644}
\vspace{-0.10cm}
\end{tabular}
\label{table:cam_ablation}
\end{adjustbox}
\vspace{-0.3cm}
\end{table}

\textbf{Module Ablation:} In our second set of ablation studies, we evaluate the effects of individual modules on the performance, namely \textit{Discretiser} and \textit{Source Encoding}.

To evaluate the importance of the \textit{Discretiser}, we devise two experiments, namely (-D-Equi) and (-D-None). In (-D-Equi), we use standard positional encoding without the \textit{Discretiser}, which means the model assumes that the consecutive frames are equidistant, \ie all the cameras share a fixed frequency. In the second setup, (-D-None), we do not provide any timing information to our \acs{aftvo}, removing both the \textit{Discretiser} and the positional encoding. To ablate \textit{Source Encoding}, (-SE), the unique one-hot vector that represents the camera sources is removed, so that the model does not know the source of its inputs.

We first share the overall performance of the proposed approach where the camera views and modules are used in Table~\ref{table:module_ablation} (See \acs{aftvo} (ours)). As can be seen, the proposed approach benefits from both the \textit{Discretiser} and the \textit{Source Encoding} modules. Removing the Discretiser, (-D-Equi), degrades the model performance drastically (resulting in nearly five times the error rate). An interesting finding is that removing the temporal information completely (-D-None) outperforms using it sequentially (-D-Equi), in the case of asynchronous data. We argue that when the data is asynchronous, using the positional encoding matrix sequentially sends the wrong time information to the fusion model. This fundamentally corrupts the temporal information, thus ultimately lowering the performance of fusion, even when compared to the case where no time information is given. Overall, this experiment shows the importance of providing accurate time information when fusing asynchronous measurements, thus validating the importance of our Discretiser module. 

\begin{table}[!h]
\centering
\caption{Average RPE (m) results from the module ablation experiments. The minus sign (-) represents that the module is ablated. SE: Source Encoding, D: Discretiser.}
\begin{tabular}{l||ccc}
Ablation & {RMSE} & {Max} & Mean $\pm$ std \\ \hline
\acs{aftvo} (ours) & \textbf{0.04529} & 0.14983 & \textbf{0.03637} $\pm$ \textbf{0.02644} \\ \hline
-D-Equi & 0.24042 & 0.82568 & 0.20731 $\pm$ 0.11833 \\
-D-None & 0.16027 & 0.56651 & 0.13559 $ \pm$ 0.08446 \\ \hline
-SE & 0.04685 & \textbf{0.14979} & 0.03731 $\pm$ 0.02778 \\
\end{tabular}
\label{table:module_ablation}
\vspace{-0.3cm}
\end{table}

\begin{table*}[!ht]
\centering
\caption{Average RPE (m) results on the nuScenes test sequences, categorised by weather/lighting conditions. NR represents where an approach failed to produce results.}
\begin{adjustbox}{width=0.95\textwidth}

\begin{tabular}{c|ccc|ccc|ccc}
\multirow{2}{*}{} & \multicolumn{3}{c|}{Daylight (64 seq.)} & \multicolumn{3}{c|}{Rain (12 seq.)} & \multicolumn{3}{c}{Night (24 seq.)} \\ \cline{2-10} 
          & RMSE & Max   & Mean $\pm$ std & RMSE & Max   & Mean $\pm$ std  & RMSE & Max  & Mean $\pm$ std \\ \hline
{ORB-SLAM \cite{mur2015orb}} & 0.398 & 3.281  & 0.120 $\pm$ 0.381 & 0.762 & 10.19 & 0.143 $\pm$ 0.736 & NR    & NR     & NR \\
{DeepVO \cite{wang2017deepvo}} & 0.086 & 0.537 & 0.057 $\pm$ 0.066 & 0.070 & 0.390 & 0.049 $\pm$ 0.049 & 0.122 & 0.742 & 0.085 $\pm$ 0.087 \\
{MDN-VO \cite{kaygusuz2021iros}} & 0.067 & 0.529 & 0.039 $\pm$ 0.053 & 0.060 & 0.434 & 0.037 $\pm$ 0.046 & 0.108 & 0.734 & 0.071 $\pm$ 0.080\\ \hline
{EKF \cite{moore2016generalized}} & 0.070 & 0.352 & 0.046 $\pm$ 0.052 & 0.065 & 0.368 & 0.044 $\pm$ 0.047 & 0.160 & 0.895 & 0.089 $\pm$ 0.129 \\
\multicolumn{1}{c|}{\acs{aftvo} (ours)} & \textbf{0.039} & \textbf{0.129} & \textbf{0.031 $\pm$ 0.023} & \textbf{0.038} & \textbf{0.133} & \textbf{0.029 $\pm$ 0.024} & \textbf{0.066} & \textbf{0.213} & \textbf{0.053 $\pm$ 0.037} \\ 
\end{tabular}
\end{adjustbox}
\label{table:nuscenes}
\vspace{-0.3cm}
\end{table*}
\begin{figure*}[!ht]
\centering
\begin{subfigure}[b]{.28\linewidth}
\centering
  \includegraphics[trim={0 0.2cm 0 0},clip, width=\textwidth]{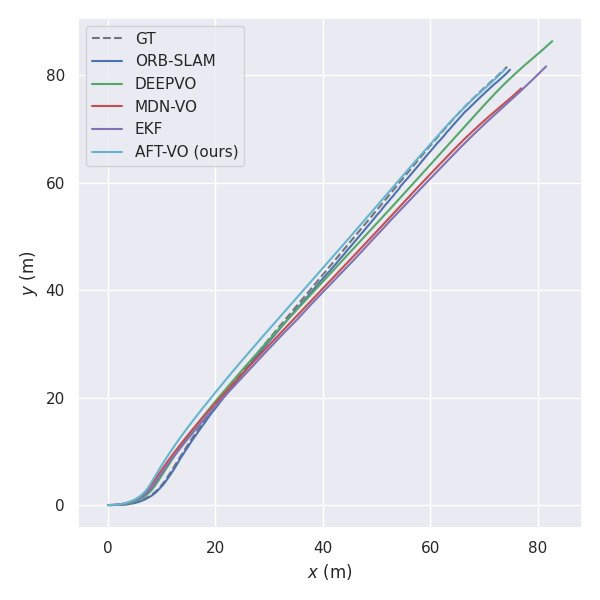}
  \caption{Daylight / scene-0016}
  \label{fig:nuscenesresult:a}
\end{subfigure}
\begin{subfigure}[b]{.28\linewidth}
\centering
  \includegraphics[trim={0 0.2cm 0 0},clip,width=\textwidth]{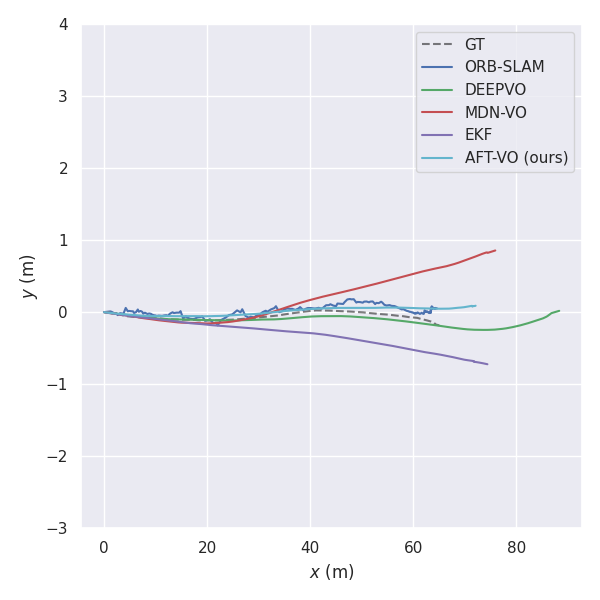}
  \caption{Rain / scene-0902}
  \label{fig:nuscenesresult:b}
\end{subfigure}
\begin{subfigure}[b]{.28\linewidth}
\centering
  \includegraphics[trim={0 0.2cm 0 0},clip,width=\textwidth]{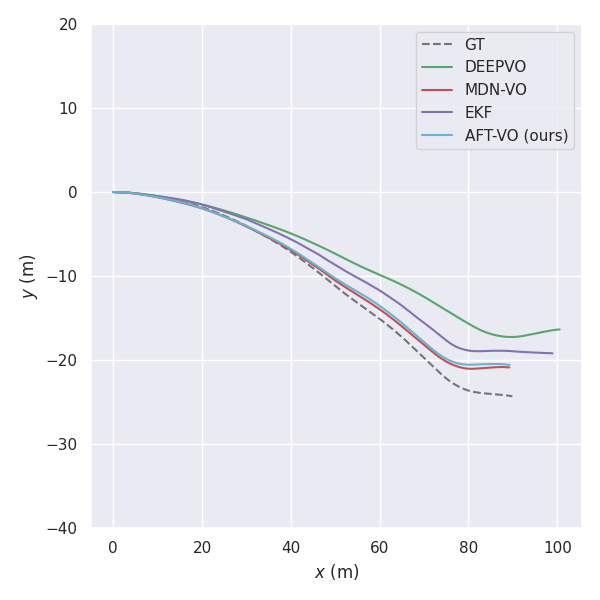}
    \caption{Night / scene-0999}
    \label{fig:nuscenesresult:c}
\end{subfigure}
\caption{Estimated \ac{vo} trajectories on nuScenes - an example from every category (Daylight, Rain, Night) is shown.}
\vspace{-0.3cm}
\label{fig:nuscenesresult}
\end{figure*}

\subsection{Comparison Against the State-of-the-Art}

In our next set of experiments, we compare the performance of our approach against state-of-the-art motion estimation techniques, namely monocular ORB-SLAM \cite{mur2015orb}, DeepVO \cite{wang2017deepvo}, MDN-VO \cite{kaygusuz2021iros} and an \ac{ekf} based fusion. 

We run ORB-SLAM with its global loop closure disabled in order to achieve a comparison in the context of \ac{vo}, as has been done previously in the literature \cite{li2018undeepvo, wang2018end}. We use the PyTorch implementation \cite{deepVOchi} of DeepVO. In order to compare our results with a classic asynchronous fusion algorithm, we employ an \ac{ekf} and use its ROS implementation \cite{moore2016generalized}. 

We use the front camera images for DeepVO, MDN-VO and ORB-SLAM which is a standard in \ac{vo} approaches. We feed the \ac{ekf} with the \ac{vo} and uncertainty estimations from the \ac{mdn} module (See Section~\ref{sec:mdn}). The reason behind this choice is to be able to compare the performance of our fusion module independently from its inputs.
Note that although AFT-VO performance is comparable to the work of Kaygusuz \etal \cite{kaygusuz2021itsc}, their approach does not account for time in an explicit manner while fusing information coming from multiple cameras. Instead, they remove asynchronicity from the data by interpolating the missing time points for predictions coming from all cameras based on a common time axis, which has the advantage of looking into the future.
Unlike their approach, our \acs{aftvo} model does not need such an interpolation step due to the proposed time encoding and hence can fuse asynchronous camera information in real-time.

\begin{table*}[!ht]
\centering
\caption{RPE (m) results on the KITTI Dataset}
\begin{adjustbox}{width=0.95\textwidth}
\begin{tabular}{c|clc|clc|clc|clc}
\multirow{2}{*}{Sequence} & \multicolumn{3}{c|}{ORB-SLAM \cite{mur2015orb}} & \multicolumn{3}{c|}{DeepVO \cite{li2018undeepvo}} & \multicolumn{3}{c|}{MDN-VO \cite{kaygusuz2021iros}} & \multicolumn{3}{c}{AFT-VO (ours)} \\ \cline{2-13} & RMSE & Max & Mean $\pm$ std & RMSE & Max  & Mean $\pm$ std & RMSE & Max & Mean $\pm$ std & RMSE & Max & Mean $\pm$ std \\ \hline
03 & \textbf{0.03} & \textbf{0.20} & \textbf{0.03 $\pm$ 0.02} & 0.08 & 0.23 & 0.08 $\pm$ 0.04 & 0.12 & 0.41 & 0.11 $\pm$ 0.06 & 0.09 & 0.29 & 0.08 $\pm$ 0.05\\
05 & 0.25 & 0.67 & 0.20 $\pm$ 0.15 & 0.24 & 0.57 & 0.21 $\pm$ 0.12   & 0.16 & 0.36 & 0.15 $\pm$ 0.08 & \textbf{0.10} &  \textbf{0.26} & \textbf{0.08 $\pm$ 0.04}\\
06 & 0.34 & 0.65 & 0.30 $\pm$ 0.17 & \textbf{0.16} & \textbf{0.31} & \textbf{0.14 $\pm$ 0.08}  & 0.20 & 0.45 & 0.18 $\pm$ 0.09 & 0.17 & 0.33 & 0.15 $\pm$ 0.08\\
07 & 0.17 & 0.35 & 0.13 $\pm$ 0.09 & 0.14  & 0.35 & 0.12 $\pm$ 0.07 & \textbf{0.08} & 0.51 & \textbf{0.07 $\pm$ 0.05} & 0.10 & \textbf{0.29} & 0.09 $\pm$ 0.04\\
10 & 0.30  & 0.95 & 0.23 $\pm$ 0.20 & 0.21 & 0.47 & 0.19 $\pm$ 0.08  & 0.14 & \textbf{0.32} & 0.13 $\pm$ 0.06 & \textbf{0.13} &  \textbf{0.32} & \textbf{0.12 $\pm$ 0.06} \\ \hline
mean & 0.22 &  0.56 & 0.18  $\pm$ 0.13 & 0.17 & 0.39 & 0.15 $\pm$ 0.08 & 0.14 & 0.41 & 0.13 $\pm$ 0.07 & \textbf{0.12} & \textbf{0.30} & \textbf{0.10 $\pm$ 0.06}\\
\end{tabular}
\end{adjustbox}
\label{table:kitti}
\vspace{-0.1cm}
\end{table*}

In our nuScenes experiments, we split the test set into three categories defined as: daylight, rain and night. The reason behind this choice is to provide a fair comparison for classical \ac{vo} approaches. For example, as a classical, feature-based approach, ORB-SLAM could not initialise on the night-time sequences, due to insufficient illumination and texture. Thus, we could not report any night-time results for ORB-SLAM.

Average \acp{rpe} on the testing sequences with respect to their categories can be seen in Table~\ref{table:nuscenes}. As a classical approach, ORB-SLAM needs to track hand-crafted features in consecutive frames. In our experiments, even though it achieves good performance in some scenarios, \ie in good weather and illumination, it loses track in others which degrades its overall performance. By contrast, DeepVO and MDN-VO achieve better performance than ORB-SLAM in all categories. The \ac{ekf} performs the second-best in daylight and rain. However, it performs worse than DeepVO and MDN-VO in night driving conditions. We believe this is due to the initial parameter selection for the Kalman Filter. We choose its initial parameters \eg process noise, initial state covariance, that achieves the best performance across all test sequences. It is worth noting that, although \ac{ekf} and our fusion approach use the \ac{mdn} outputs (see Section~\ref{sec:mdn}), our fusion module performs better than the \ac{ekf}. As can be seen, our proposed method, \acs{aftvo}, achieves the best performance in all three categories which validates our transformer-based fusion approach to estimate \ac{vo}.

We demonstrate an example of estimated \ac{vo} trajectories from three different categories in Figure~\ref{fig:nuscenesresult}. It can be seen that the \ac{vo} trajectories estimated by our approach, \acs{aftvo}, are generally more accurate than the other approaches. Although ORB-SLAM's performance in daylight is successful, it tends to lose track in the rainy scenario which decreases its performance. As a learning-based approach, DeepVO and MDN-VO are able to produce results even for night scenes.

Results on the KITTI dataset can be seen in Table~\ref{table:kitti}. 
As with the nuScenes dataset, \acs{aftvo} achieves the best overall mean performance across all sequences. 
Even though ORB-SLAM achieves the best performance on Sequence 03, it has larger error on other sequences. In general, DeepVO and MDN-VO results are better than ORB-SLAM. However, \acs{aftvo} achieves the best results on most of the sequences. It is worth noting that, even though KITTI has two synchronised cameras, our fusion approach achieves the best results on most sequences further validating the proposed transformer-based fusion approach.

\section{Conclusion}
\label{sec:conclusion}
In this paper, we presented a novel transformer-based fusion architecture to estimate \ac{vo}. We formalised \ac{vo} fusion as a sequence-to-sequence problem and employ transformers to fuse measurements from multiple asynchronous cameras. We proposed a novel \textit{Discretiser} method which enables the network to positionally encode continuous timestamps, allowing transformer architectures to seamlessly handle asynchronous data. We also proposed \textit{Source Encoding} to enable our fusion module to differentiate the source of measurements that it is getting. We evaluated \acs{aftvo} on two popular datasets, namely nuScenes and KITTI, and reported state-of-the-art performance. As future work, we are planning to expand our asynchronous fusion approach by including other sensor modalities such as IMU.

\section{Acknowledgements}
\label{sec:Acknowledgements}
This paper was supported by the EPSRC project ROSSINI (EP/S016317/1).

{\small
\bibliographystyle{ieee_fullname}
\bibliography{egbib}
}

\end{document}